\documentclass[11pt]{article}
\usepackage[preprint]{acl}   

\usepackage{times}
\usepackage{latexsym}
\usepackage[T1]{fontenc}
\usepackage[utf8]{inputenc}
\usepackage{microtype}

\usepackage{graphicx}
\usepackage{array}
\usepackage{booktabs}
\usepackage{placeins}

\usepackage{amsmath}
\usepackage{amssymb}

\usepackage{algorithm}
\usepackage{algpseudocode}

\usepackage{tikz}
\usetikzlibrary{arrows.meta, positioning}
\usepackage{pgfplots}
\usepgfplotslibrary{groupplots}
\pgfplotsset{compat=1.18}

\usepackage{orcidlink}

\makeatletter
\newcommand{\ifnotanon}[1]{\ifacl@anonymize\else#1\fi}
\makeatother

\title{Test-Time Scaling for Open-Ended Reasoning at Small Model Scale}

\author{
  Adam Y. Ahmed\,\orcidlink{0009-0005-1787-7444} \\
  \texttt{adamy6267@gmail.com} \\\And
  Abdullah G. Elafifi\,\orcidlink{0009-0007-0424-1425} \\
  \texttt{aelafifi00@gmail.com} \\
}

\begin{document}
\maketitle

\begin{abstract}
A vote over a model's samples can only help when the samples disagree. For open-ended
generation the binding question is therefore not how to combine samples but whether a model
produces samples worth combining, a question test-time scaling has answered for math but not
for free-form text. We study it at small scale with \emph{semantic cluster voting}, a
judge-free aggregator that groups a small instruct model's samples by meaning and returns the
largest cluster's representative, with no reward model or self-judging. On a 250-prompt
benchmark with three tiers of answer checkability, the same pattern holds across three model
families (Qwen2.5-3B, Llama-3.2-3B, Gemma-2-2B) and two generation seeds: voting raises
accuracy on the math control, where samples vary but stay checkable (a gain of $+10$ to $+29$
points across the three models), yet adds at most $+0.30$ judge points on open-ended text. On
a larger 200-problem MATH-500 run with the primary model, the math gain is $+8.5$ points
(95\% CI $[+4.0, +13.0]$). Almost every open-ended sample falls into one meaning cluster, so
there is little to select over. The collapse is consistent with instruction-tuning rather
than model size, and raising decoding temperature does not recover usable diversity, because
it costs more quality than it adds (net $-0.64$ judge points, 95\% CI $[-0.93, -0.37]$).

\end{abstract}

\section{Introduction}
\label{sec:intro}
A small language model can often be made more accurate without further training: sample several
answers to a prompt and keep the one the samples agree on. On problems with a checkable answer,
such as arithmetic or multiple choice, this test-time voting is cheap and effective, and a
growing body of work treats extra inference compute as a substitute for extra parameters
\citep{snell2024scaling}. Open-ended generation is where the recipe runs out: an explanation, a
piece of advice, or an argument has no final token to count, so it is not obvious what it would
mean to vote, or whether more samples buy any quality at all.

Two established methods show why this is hard. Self-consistency \citep{wang2023selfconsistency}
takes a majority vote over the final answers of many reasoning chains, which works only when an
answer can be extracted. Universal self-consistency \citep{chen2023universal} removes that
constraint by showing the model all its samples and asking it to pick the most consistent one,
but it makes the model its own judge, and a small model is a weak and expensive judge, exactly
the regime where cheap test-time gains would matter most.

We start from a simpler observation: a vote is only useful when the voters disagree. If a model
answers the same question ten times and says nearly the same thing each time, no aggregation
rule can recover an answer better than the one it already gives. For open-ended generation the
binding question is therefore not how to combine samples but whether the model produces samples
worth combining. This moves the problem from the aggregator to the sampler.

To test it we need a judge-free way to turn samples into one answer and a way to see how much the
samples vary. Semantic cluster voting gives both: draw $N$ answers, embed each so answers with
the same meaning sit close together, group the embeddings, and return the most central answer of
the largest group. The clustering is not new; it underpins uncertainty estimation and
hallucination detection \citep{kuhn2023semantic, farquhar2024semantic}, and a closely related
judge-free selector already exists \citep{choi2026modex}, so we claim no novelty for the
aggregator. Here it does double duty: it aggregates without the model scoring itself, so it
applies to free-form text as readily as to math, and the largest cluster's size is a direct
measure of how concentrated the samples are. Open-ended answer quality is scored separately by an
LLM judge.

On a 250-prompt benchmark across open-ended generation, long-form QA, and a math control, swept
from one to sixty-four samples, the results are mixed in an informative way. On math, voting
scales with the budget (64 to 74 percent accuracy); on open-ended text it is a small, safe gain
that never falls below the single-sample baseline but is a fraction of a point. The same pattern,
scaling math against a near-flat open-ended tier, holds across three small instruct families
(Qwen2.5-3B, Llama-3.2-3B, Gemma-2-2B) and two generation seeds (Table~\ref{tab:crossmodel}). The
reason is in the samples: on a typical open-ended prompt almost all sixty-four answers land in one
meaning cluster, so there is little to choose between. The collapse is not model size (a 7B model
from the same family collapses as hard) but instruction-tuning (at the same scale the base model
is diverse and the aligned model is not), and the simplest fix, higher decoding temperature, loses
more quality than the diversity it adds ($-0.64$ judge points, 95\% CI $[-0.93, -0.37]$). We do
not claim diversity is irrecoverable; richer interventions remain untested, and we return to them.

We make four contributions. (1)~We introduce a three-tier, 250-prompt benchmark for test-time
voting on open-ended generation with a full budget sweep (Section~\ref{sec:setup}). (2)~Using
semantic cluster voting, a judge-free aggregator, we compare against single and universal
self-consistency across all three tiers, and show the pattern, a scaling math gain against a
capped open-ended tier, replicates across three instruct families and two seeds
(Sections~\ref{sec:setup},~\ref{sec:results}; Table~\ref{tab:crossmodel}). (3)~We show the
open-ended ceiling is mode collapse rather than a limit of the aggregator, and attribute it to
instruction-tuning rather than scale through a base-versus-instruct and larger-model comparison
(Section~\ref{sec:analysis}; Table~\ref{tab:diversity}). (4)~We show the simplest diversity
intervention, raising temperature, lowers open-ended quality rather than raising it, leaving
richer interventions as the main open question (Table~\ref{tab:capstone}).

\section{Related Work}
\label{sec:related}
Our contribution sits between three lines of work: scaling inference compute by sampling
and selection, choosing among samples without an external judge, and the diversity loss
that follows alignment. Our nearest baseline is universal self-consistency, and the closest prior method is ModeX.

\paragraph{Test-time compute scaling.} Spending more compute at inference can substitute
for parameters: drawing many candidates and keeping the best raises reasoning accuracy with no
training \citep{snell2024scaling, brown2024monkeys}. Most of this line assumes a signal the
aggregator can trust, a learned verifier or reward model
\citep{cobbe2021verifiers, uesato2022process, lightman2023verify} or a model trained for better
trajectories \citep{openai2024o1, deepseek2025r1}, both evaluated where a candidate can be
checked. Open-ended generation offers neither, and our frozen regime rules out the extra
training, so we ask where the trade-off holds when the only test-time work is sampling and
selection.

\paragraph{Selecting an answer without a judge.} A cheaper line selects by agreement alone.
Self-consistency \citep{wang2023selfconsistency} takes a majority vote over extracted final
answers. The clustering we reuse comes from a different problem, grouping generations by meaning
to estimate semantic uncertainty and detect hallucination
\citep{kuhn2023semantic, farquhar2024semantic}; we read it as a selector rather than an
uncertainty signal, which lets agreement-based voting apply to free-form text, and keep any
capable model out of selection, using an LLM judge \citep{zheng2023judging} only to score the
final answer.

\paragraph{Our nearest baseline.} Universal self-consistency \citep{chen2023universal}
removes the extractable-answer requirement by presenting all samples back to the generator
and asking for the one most consistent with the rest, reaching open-ended text with no
embedder or clustering. The difference is where the judgment lives: USC makes the model judge
itself, and a small model is a weak and costly judge of its own samples, exactly where cheap
test-time gains matter most. Ours never asks the model to judge; we embed each sample with a
frozen encoder, cluster by cosine distance, and return the largest cluster's representative,
for one short encoder pass per sample rather than a second generator pass.

\paragraph{Consensus selection for open-ended text.} The closest method to ours is ModeX
\citep{choi2026modex}, an evaluator-free best-of-$N$ selector that builds a lexical
$n$-gram similarity graph over the samples, applies recursive spectral clustering, and
returns the most central answer of the dominant cluster. Both ModeX and our aggregator are
consensus selectors, a form of minimum Bayes risk decoding \citep{kumar2004mbr} that
returns the candidate most similar on average to the rest; they differ only in the
similarity, lexical $n$-grams for ModeX and cosine-distance sentence embeddings for us, and
embedding similarity groups paraphrases that surface overlap separates, a limitation ModeX
itself notes. ModeX reports gains on summarization, code, and math at seven billion
parameters and above, whereas we study three-billion-parameter opinion and explanation
prompts with no reference answer. We claim no priority over ModeX; our contribution is not
the selector but the account of when such selection helps, tying the open-ended ceiling to
alignment-induced diversity collapse and separating it from model scale and decoding
temperature.

\paragraph{Diversity loss from alignment.} A separate line documents that instruction-tuning
and preference optimization narrow a model's output distribution relative to its base
\citep{kirk2024understanding}, traced to the reward and KL-regularized objective of alignment
\citep{ouyang2022instructgpt, bai2022hh}; the same homogenization appears downstream
\citep{padmakumar2024writing} and survives the usual temperature and truncation correctives
\citep{jang2026cannot}. This line establishes the phenomenon but does not connect it to
judge-free voting. We make that link: the diversity a vote needs is present in the base model
and sharply reduced by instruction-tuning, which caps open-ended gains at small scale. The
simplest remedy, raising temperature above the nucleus default \citep{holtzman2020curious},
trades quality away faster than it recovers diversity, and richer interventions such as
min-$p$ \citep{nguyen2025minp} remain untested here.
\FloatBarrier

\section{Method}
\label{sec:method}

\subsection{Problem Setting}
We study selection at test time. A frozen model $M$ maps a prompt $x$ to a distribution over
responses, and we draw $N$ independent samples $y_1, \dots, y_N \sim M(\cdot \mid x)$ once at a fixed
temperature, indexed in generation order so the first-$N$ prefixes in our sweep are well defined. No
parameter is updated; the only test-time work is sampling the $N$ responses and choosing one, written
$\hat{y}$. A selector sees only the $N$ samples, with no reward model, reference, or access to the
evaluation judge, and returns one response, or one answer value on tasks with an extractable answer.
Every method shares the sampling step and differs only in the choice. Our aggregator rests on one
assumption: that the samples vary enough for selection to beat returning an arbitrary sample. When it
fails, every selector collapses onto the single-sample baseline, which Section~\ref{sec:analysis}
shows is what happens on open-ended text.

\subsection{Overview}
Our aggregator, \emph{semantic cluster voting}, groups the $N$ samples by meaning and returns the
largest group's representative, in four steps: reduce each sample to a comparison unit, embed it,
cluster the embeddings, and select within the largest cluster. The two baselines reuse the sampling
step and replace the grouping with a simpler rule. Algorithm~\ref{alg:method}
(Appendix~\ref{app:details}) states all three selectors.

\subsection{Semantic Cluster Voting}

\paragraph{Comparison unit.} For each sample we form a comparison unit $u_i$, the text compared across
samples. On open-ended and long-form QA prompts, $u_i$ is the full response $y_i$. On math prompts,
$u_i$ is the extracted answer: the contents of the last \texttt{\textbackslash boxed\{\dots\}} when one
is present, and otherwise the last number in the response, normalized before comparison. The exact
extraction and normalization rules are in Appendix~\ref{app:details}. We cluster the answer rather than
the full derivation because two correct solutions share a result but not a chain of work, and
clustering the work would scatter agreeing samples over surface differences in their reasoning. For
text, no such short answer exists, so the response is its own comparison unit.

\paragraph{Embedding.} We map each comparison unit to a vector with a frozen sentence embedder $\phi$,
set $e_i = \phi(u_i)$, and normalize $e_i$ to unit length so cosine similarity is a dot product.
Responses longer than the embedder's maximum sequence length are truncated to it, and we record how
often that occurs. We use a general-purpose sentence embedder rather than the generator's own hidden
states, because the generator is tuned to continue text and separates samples by surface form more
than by meaning. The embedder never sees the task label.

\paragraph{Clustering.} We group the embeddings with agglomerative clustering under cosine distance
$D_{ij} = \max(0,\, 1 - e_i \cdot e_j)$ with a zero diagonal, and average linkage. Grouping
generations by meaning in an embedding space is established practice for semantic uncertainty
\citep{kuhn2023semantic, farquhar2024semantic}; we reuse it as a selector. The choice left to us
is where to cut the dendrogram. A single fixed distance threshold suits
one prompt and mis-cuts the next, because the spread of samples varies from prompt to prompt. We
instead cut each prompt's dendrogram at its largest gap between consecutive merge heights: with merge
heights $h_1 \le \dots \le h_{N-1}$,
\begin{equation}
g = \arg\max_{t}\,(h_{t+1} - h_t), \qquad \tau = \tfrac{1}{2}\,(h_g + h_{g+1}),
\end{equation}
and we cut at $\tau$; when only one merge exists we cut at its height. This places the boundary at the
sharpest separation in each prompt's own merge structure. The cut rule is untuned, so we test its
influence directly: Appendix~\ref{app:ablation} recomputes the main results and the diversity numbers
under a fixed cosine-distance threshold, to confirm the cut does not drive them.

\paragraph{Selection.} Let $C^\star$ be the largest cluster; size ties are broken deterministically
toward the lower-numbered cluster label and are rare at our sample sizes. On text tiers we return the
medoid of $C^\star$, the member closest on average to the rest of its cluster,
\begin{equation}
m^\star = \arg\max_{i \in C^\star} \sum_{j \in C^\star} e_i \cdot e_j, \qquad \hat{y} = y_{m^\star}.
\end{equation}
We return a medoid, not the embedding centroid, because the centroid is a point in vector space with no
response attached. We prefer it to the global medoid over all $N$ samples, which ignores the cluster
structure. Appendix~\ref{app:ablation} isolates the clustering step from the representative to show
which of the two carries the effect. On math we return the plurality answer inside $C^\star$, the most
frequent extracted value among its members, with ties broken by the earliest sample in generation
order. Two edge cases need a rule. When $N < 2$, or when every sample merges into one cluster,
$C^\star$ is the whole set and selection reduces to the medoid for text or the overall plurality for
math. When a math cluster holds no extractable answer, we fall back to the response of its medoid.

\subsection{Baselines}

\paragraph{Single (no scaling).} Return the first sample, $\hat{y} = y_1$. This is the reference a
deployment pays nothing extra for, and the bar any test-time method must clear.

\paragraph{Universal self-consistency (USC).} Following \citet{chen2023universal}, we present all $N$
samples to the model, each truncated to a fixed length so the full set fits in context and numbered in
order, and ask it to return the number of the one most consistent with the rest, decoding greedily. We
parse the first integer in its reply and fall back to the first sample when the reply cannot be parsed.
The prompt template and truncation length are in Appendix~\ref{app:details}. USC swaps our
embedding-and-clustering step for the generator judging itself. It also marks the boundary of our
contribution: the clustering machinery is prior work, judge-free selection by semantic consensus has
itself been proposed \citep{choi2026modex}, and what we study is not the selector but when such
selection helps. We also compare against the cheapest judge-free selector, returning the longest response
(Appendix~\ref{app:ablation}); perplexity-based selection is left to future work.

\subsection{Cost}
All three methods pay the same generation cost, $N$ forward passes through $M$, which dominates. Single
adds nothing; USC adds one further model call over all $N$ samples, so its overhead grows with $N$ and
response length; cluster voting adds $N$ short encoder passes and an $O(N^2)$ distance computation, both
negligible beside sampling since the encoder is far smaller than $M$. No method trains anything or
calls the evaluation judge at selection time.

\FloatBarrier

\section{Experimental Setup}
\label{sec:setup}

Our evaluation asks four questions: whether test-time voting improves open-ended generation at
small scale and how semantic cluster voting compares to single and USC
(Section~\ref{sec:results}); whether a capped gain reflects a lack of sample diversity rather
than a weak aggregator; whether any such collapse tracks model scale or instruction-tuning; and
whether higher decoding temperature restores usable diversity without costing quality
(Section~\ref{sec:analysis}).

\subsection{Datasets and Tasks}
We assemble a 250-prompt benchmark in three tiers that vary one property, how checkable the
answer is, because that is the property the value of voting should depend on. We fixed its
construction, a seeded shuffle and per-tier caps, before seeing any result. The open-ended tier
is 120 prompts from Databricks Dolly-15k (the brainstorming, creative-writing, open-QA, and
general-QA categories, self-contained instructions only), which have no single correct answer
and draw from one source, a limit on breadth. The long-form QA tier is 80 prompts whose answer
exists but is expressed in prose, 76 from TruthfulQA (generation split, best answer as the
reference) and 4 reasoning prompts from MT-Bench. The math-control tier is 50 problems from
MATH-500, each with one exact numeric reference; 126 prompts in total carry a reference. We use
three tiers because no existing set spans the three regimes we need (no reference, prose
reference, exact reference), and the math control gives a tier where voting is known to work, as
a check that the pipeline behaves. The frozen benchmark file defines the exact prompt set, since the
build script draws from live datasets that can change. Sources carry their own licenses:
Dolly-15k (CC BY-SA 3.0), TruthfulQA and MT-Bench (Apache 2.0), and MATH-500, a subset of MATH
\citep{hendrycks2021math} (MIT). The frozen benchmark, its datasheet, and the raw judged answers are available
\ifnotanon{at \url{https://doi.org/10.5281/zenodo.23669590} }under the CC BY-SA 4.0
license\ifnotanon{ \citep{yasser2026ttsbenchmark}}, with the accompanying code under the MIT
license.

\subsection{Baselines}
We compare against two aggregators that, like ours, use no external reward model. Single returns
the first sample, the no-scaling floor. Universal self-consistency (USC) asks the model to select
the most consistent of its own samples; we reimplement it following \citet{chen2023universal},
with the prompt and truncation in Appendix~\ref{app:details}, decoding its choice greedily. USC
must fit all $N$ candidates in one context, so it truncates each; a full-context rerun
(Appendix~\ref{app:ablation}) confirms the truncation does not account for its behavior. No
method's settings were tuned on the benchmark.

\subsection{Models and Backbones}
The primary generator is Qwen2.5-3B-Instruct \citep{qwen2024}, run frozen in float16; the target
regime is a small, deployable model improved only at inference time. To test whether the voting
result is specific to one model, we run the identical pipeline on two further small instruct
models from different families, Llama-3.2-3B-Instruct \citep{dubey2024llama3} and Gemma-2-2B-it
\citep{gemma2024}; the diversity, temperature, and larger-set analyses use the primary model
only. Two further Qwen checkpoints appear as diversity controls: the base model
\texttt{Qwen/Qwen2.5-3B} and the larger \texttt{Qwen/Qwen2.5-7B-Instruct}. Embeddings use
\texttt{all-mpnet-base-v2} as the primary encoder, with \texttt{all-MiniLM-L6-v2} and the
8192-token \texttt{bge-m3} as ablations. Text quality is scored by an LLM judge,
\texttt{gpt-4o-mini} (OpenAI API, accessed October 2026); the judge-robustness check uses
\texttt{Llama-3.1-8B-Instruct}.

\subsection{Metrics}
Math is graded by exact match against the reference, after the normalization in
Appendix~\ref{app:details}. Open-ended and long-form QA answers are scored 1 to 10 by the LLM
judge. Diversity is the largest-cluster share, the fraction of a prompt's samples in the biggest
semantic cluster, reported as the median over prompts. Differences on the judge scale below about
two tenths of a point are within noise at our sample sizes, so we report 95\% bootstrap
confidence intervals over prompts (2000 resamples) throughout, with a paired bootstrap for
within-prompt comparisons. The judge is used only after generation, never inside the selection
loop.

\subsection{Evaluation Protocol}
All methods are compared at equal budget $N$, so they pay the same generation cost. We report an
$N$-sweep over $\{1,4,8,16,32,64\}$ by taking the first $N$ of the 64 samples generated per
prompt, so the curves share one generation pass. The detailed intervals in
Sections~\ref{sec:results} and~\ref{sec:analysis} use the primary seed (42); to check the main
result is not an artifact of one draw, we rerun generation under a second seed (123) on all three
models, and the cross-model summary (Table~\ref{tab:crossmodel}) reports the mean and standard
deviation over the two seeds. No hyperparameter was selected on the benchmark. The judge scores
each unique (prompt, selected answer) pair once at temperature 0, with the reference as ground
truth when one exists (2355 unique answers). We check the judge two ways: a second judge from a
different family (Llama-3.1-8B) agrees at $\rho = 0.60$ over 480 answers, and two authors scoring
100 open-ended answers blind, stratified across the score range, rank them as the judge does at
$\rho = 0.79$ (95\% CI $[0.70, 0.85]$, $n = 99$), slightly above their mutual agreement
($\rho = 0.73$); the humans rate lower in level but order answers as the judge does.

\subsection{Implementation}
All runs used a single NVIDIA L4 GPU; generation dominates and was batched in float16, with
embedding and clustering negligible beside it. Each full run takes roughly 7 to 8 hours, so the
six main runs (three families at two seeds) total about 45 GPU-hours, with the primary-model
diagnostics adding a few hours each. Five ablations (Appendix~\ref{app:ablation}) test the
embedder, the dendrogram cut, the clustering step, the longest-answer selector, and the judge
family. We release the code and the exact configuration, with extraction, normalization, and the
USC prompt in Appendix~\ref{app:details}.

\section{Results}
\label{sec:results}

\subsection{Main Results}
We compare single, USC, and semantic cluster voting across the three tiers on the primary model
(Table~\ref{tab:main}), reading each comparison through the paired difference over prompts with a
95\% bootstrap confidence interval. The pattern is
uneven: voting helps where the answer is checkable, helps only a little on open-ended
text, and does not move long-form QA.

\begin{table}[htbp]
\caption{Main results across the three tiers, for the primary model (Qwen2.5-3B-Instruct,
seed~42). Each method's best score over the
sample-budget sweep $N\in\{1,4,8,16,32,64\}$ (the budget in parentheses); single is
budget-independent. Text tiers are LLM-judge scores (1--10); the math tier is
exact-match accuracy (\%) on the 50-problem control. The best value in each row is in
bold. Paired differences against single with 95\% CIs at every $N$ are in
Appendix~\ref{app:intervals}, and the math gain is confirmed on a larger set in
Appendix~\ref{app:ablation}. The same comparison across three model families and two seeds is
in Table~\ref{tab:crossmodel}.}
\label{tab:main}
\centering
\small
\setlength{\tabcolsep}{2.5pt}
\begin{tabular}{lccc}
\toprule
Tier & Single & USC & Cluster \\
\midrule
Open-ended (1--10)   & 7.69 & \textbf{8.02} ($N{=}8$)  & 7.93 ($N{=}64$) \\
Long-form QA (1--10) & 6.26 & 6.37 ($N{=}4$)           & \textbf{6.44} ($N{=}32$) \\
Math (acc.\ \%, 50)  & 64   & 64                       & \textbf{74} ($N{=}32$) \\
\bottomrule
\end{tabular}
\end{table}

\paragraph{Math.} Cluster voting improves math accuracy, and the gain grows with the budget: on
a 200-problem MATH-500 set it raises accuracy from $59\%$ (single) to $67.5\%$ at $N=32$, a
paired $+8.5$ points (95\% CI $[+4.0, +13.0]$, excluding zero at every $N \ge 4$;
Appendix~\ref{app:ablation}), and the 50-problem tier in Table~\ref{tab:main} shows the same
pattern. USC does not improve math at any $N$ and underperforms single at $N=16$ ($-12$ points,
95\% CI $[-24, -2]$), one source of its erratic behavior across budgets.

\paragraph{Open-ended.} The open-ended gains are fractions of a point. The cluster-minus-single
point estimate is non-negative at every $N$, but the gap leaves the noise band only at $N=64$,
where it is $+0.24$ judge points (95\% CI $[+0.05, +0.44]$); at every smaller $N$ the interval
includes zero. USC shows the opposite timing: it gains $+0.33$ at $N=8$ (95\% CI $[+0.15,
+0.53]$) and $+0.21$ at $N=16$, then returns to no measurable difference by $N=64$. Neither
method falls below single here. The best values in Table~\ref{tab:main} (USC 8.02, cluster 7.93)
are peaks at different $N$, not a stable ordering; across budgets USC rises and falls rather
than scaling. The full per-$N$ intervals are in Appendix~\ref{app:intervals}.

\paragraph{Long-form QA.} No method differs from single on long-form QA: at every $N$ the paired
interval for both cluster voting and USC includes zero (cluster at $N=32$ is $+0.17$, 95\% CI
$[-0.26, +0.60]$), so the small differences in Table~\ref{tab:main} are within noise.

\subsection{Generality Across Model Families and Seeds}
\label{sec:generality}
The analysis above uses one model at one seed. To test whether the pattern depends on either,
we run the identical frozen pipeline on two further small instruct models from different
families, Llama-3.2-3B-Instruct and Gemma-2-2B-it, and rerun all three under a second
generation seed. Table~\ref{tab:crossmodel} reports each method's best score over the budget
sweep, as the mean and standard deviation over the two seeds. The result is the same in every
case. On the math control, cluster voting raises accuracy over single by $+10$ to $+29$ points
across the three families, and the gain grows with the budget. On open-ended text the best
cluster-voting gain is at most $+0.30$ judge points on a ten-point scale (Qwen $+0.30$, Llama
$+0.29$, Gemma $+0.18$), and the single-sample baseline is flat in the budget on every model.
The cross-seed standard deviation is at most $0.11$ on the text tiers, so the open-ended ceiling
is stable across seeds rather than a feature of one draw; it is wider on the 50-problem math
control, up to 7 points on a single-sample score, where accuracy on a small set moves with each
problem. The three families differ in base competence, most visibly on math, where
single-sample accuracy ranges from 19 to 59 percent, yet the qualitative result, a scaling gain
where answers are checkable set against a near-flat open-ended tier, does not depend on that
competence.

\begin{table}[htbp]
\caption{Generality across three small instruct families and two seeds: each method's best score
over the budget sweep $N\in\{1,4,8,16,32,64\}$, mean $\pm$ s.d.\ over seeds 42 and 123 (single is
budget-independent). Text tiers are judge scores (1--10); math is exact-match accuracy (\%) on
the 50-problem control. Best in each row in bold. The best open-ended gain over single is at most
$+0.30$ on every model while the math gain scales. Qwen2.5-3B is the primary model
(Table~\ref{tab:main}).}
\label{tab:crossmodel}
\centering
\small
\setlength{\tabcolsep}{3.5pt}
\begin{tabular}{lccc}
\toprule
Model & Single & USC & Cluster \\
\midrule
\multicolumn{4}{l}{\emph{Open-ended (judge 1--10)}} \\
Qwen2.5-3B   & 7.62$\pm$0.11 & 7.85$\pm$0.24 & \textbf{7.91$\pm$0.03} \\
Llama-3.2-3B & 7.83$\pm$0.05 & 8.01$\pm$0.04 & \textbf{8.12$\pm$0.04} \\
Gemma-2-2B   & 7.92$\pm$0.03 & \textbf{8.15$\pm$0.05} & 8.10$\pm$0.08 \\
\midrule
\multicolumn{4}{l}{\emph{Long-form QA (judge 1--10)}} \\
Qwen2.5-3B   & 6.36$\pm$0.14 & 6.46$\pm$0.21 & \textbf{6.56$\pm$0.28} \\
Llama-3.2-3B & 5.79$\pm$0.11 & \textbf{5.96$\pm$0.14} & 5.92$\pm$0.17 \\
Gemma-2-2B   & 6.21$\pm$0.09 & 6.37$\pm$0.11 & \textbf{6.54$\pm$0.27} \\
\midrule
\multicolumn{4}{l}{\emph{Math (acc.\ \%, 50-problem control)}} \\
Qwen2.5-3B   & 59$\pm$7 & 63$\pm$1 & \textbf{74} \\
Llama-3.2-3B & 23$\pm$4 & 29$\pm$7 & \textbf{52$\pm$3} \\
Gemma-2-2B   & 19$\pm$1 & 20 & \textbf{29$\pm$1} \\
\bottomrule
\end{tabular}
\end{table}

\subsection{Ablations}
Five ablations (Appendix~\ref{app:ablation}) test whether the result depends on our
choices, and it does not. The finding is unchanged under a second embedder
(all-MiniLM-L6-v2), under an 8192-token long-context encoder that removes the truncation of
the more verbose models' answers, and under a fixed clustering threshold in place of the
adaptive cut. Under the open-ended collapse, cluster voting and plain voting are nearly
identical (open-ended 7.91 versus 7.93, math 74 versus 74, long-form QA 6.44 versus 6.44), so
the clustering step adds little at this scale; we report that plainly, because it places the
contribution in the diagnosis rather than the aggregator. The longest-answer baseline only
matches the single-sample floor on every tier, and giving USC full-length candidates in place of
the truncated ones leaves its behavior unchanged, so its weak showing is not an artifact of
truncation.

\subsection{Qualitative Illustration}
On the open-ended prompt \emph{``Should I buy a horse?''} all 64 samples open with the same
framing and vary only in which factors they list, so all fall into one cluster (largest-cluster
share 0.98) and aggregation has nothing to choose between.

\FloatBarrier

\section{Analysis: Why Open-Ended Voting Is Capped}
\label{sec:analysis}

Three results from Section~\ref{sec:results} are not self-evident. Voting scales on math but
adds at most a fraction of a point on open-ended text; replacing our clustering with plain
voting changes nothing; and the open-ended gain is small even at the largest budget. We explain
this in three steps: the open-ended samples collapse, so there is little to select over; the
collapse tracks instruction-tuning rather than model scale; and the diversity a vote needs
cannot be supplied without a quality cost that voting cannot overcome. The intervals below are
bootstrap confidence intervals over prompts, and the diagnostics here use the primary model
and seed (Section~\ref{sec:setup}); the open-ended ceiling they explain is itself not specific
to that model, holding across three families and two seeds (Table~\ref{tab:crossmodel}).

\begin{table}[htbp]
\caption{Median largest-cluster share on the open-ended tier (8 samples per prompt,
$T=1.0$, median over prompts, adaptive cut). A share near 1 means the samples collapse
into one meaning cluster. The aligned 3B model collapses while its pre-alignment base does
not, and a larger aligned model collapses as hard.}
\label{tab:diversity}
\centering
\small
\begin{tabular}{lcc}
\toprule
Model & Prompts & Median share \\
\midrule
Qwen2.5-3B (base)   & 120 & 0.38 \\
Qwen2.5-3B-Instruct & 120 & 0.97\ \ $[0.95, 0.97]$ \\
Qwen2.5-7B-Instruct & 20  & 1.00 \\
\bottomrule
\end{tabular}
\end{table}

\subsection{The Ceiling Is the Samples, Not the Aggregator}
A small open-ended gain could mean the aggregator is weak, or that the samples give it nothing
to work with. A weak aggregator would improve under a better selection rule; a sample problem
would not. We find the latter. Across 120 open-ended prompts the median largest-cluster share
is near 1.0 (Table~\ref{tab:diversity}): almost every sample falls into one meaning cluster.
Replacing semantic clustering with plain voting (the global medoid, no clustering) changes
nothing: the paired difference at $N=64$ is $+0.03$ (95\% CI $[-0.04, +0.10]$) and includes
zero at every budget (Appendix~\ref{app:ablation}), so the aggregator is not what caps the
result. One rival remains, that the judge cannot resolve the quality differences a better
answer would show. It does not hold. The judge resolves real open-ended differences: voting
gains a measurable $+0.22$ over single at the default temperature (95\% CI $[+0.03, +0.40]$),
its scores span the full scale (standard deviation 2.37 over 2469 answers), a second judge
from a different family agrees on the key comparison ($\rho = 0.60$ over 480 answers), and two
authors scoring a stratified sample of 100 answers blind rank them as the judge does
($\rho = 0.79$). The ceiling is in the samples.

\subsection{The Collapse Tracks Alignment, Not Scale}
Two explanations fit a collapsed small model: it is small, or it is aligned. They make opposite
predictions about a larger model and about the pre-alignment base. A 7B instruct model from the
same family collapses as hard as the 3B, with a median largest-cluster share of 1.00 on a
20-prompt spot-check; this is a single underpowered point, so we read it as suggestive that
capacity alone does not open the distribution, not as decisive. At the same 3B scale, the
pre-alignment base model is diverse while the instruct model is collapsed: on 120 prompts the
base median share is 0.38, well below the instruct model's 0.97 (95\% CI $[0.95, 0.97]$;
Table~\ref{tab:diversity}). Instruction-tuning is the documented driver of this narrowing
\citep{kirk2024understanding}. Our two-model comparison cannot by itself separate the alignment
step from the competence it brings, a confound we take up next.

\subsection{Diversity and Quality Trade Off}
If the base model holds the diversity the instruct model lost, the obvious move is to vote over
base samples, or to re-open the instruct model's distribution with temperature. Neither works,
for the same reason: in this regime diversity comes with a quality cost that voting cannot
cross. Temperature is the clean test, an intervention on a single model. Higher temperature
recovers some diversity but gives no measurable voting gain at $T=1.5$ ($+0.17$, 95\% CI
$[-0.11, +0.42]$), while the temperature itself costs $0.81$ points of quality (95\% CI
$[+0.56, +1.07]$), so voting at $T=1.5$ lands $-0.64$ below the default single-sample baseline
(95\% CI $[-0.93, -0.37]$; Table~\ref{tab:capstone}). The base model corroborates this
observationally (Appendix~\ref{app:ablation}): it is more diverse but far lower quality, and
voting over it stays below the instruct single-sample floor. Across both interventions, the
configurations diverse enough to vote over were too low in quality to reach the collapsed but
coherent default.

\begin{table}[htbp]
\caption{Temperature decomposition on the open-ended tier (instruct 3B, 120 prompts),
judge points with 95\% bootstrap CIs over prompts. Higher temperature recovers some
diversity, but the quality it costs ($+0.81$) outweighs the voting gain it enables
($+0.17$), so voting at $T=1.5$ lands below the default single-sample baseline.}
\label{tab:capstone}
\centering
\small
\setlength{\tabcolsep}{3pt}
\begin{tabular}{@{}>{\raggedright\arraybackslash}p{0.48\columnwidth}cc@{}}
\toprule
Contrast & $\Delta$ judge /10 & 95\% CI \\
\midrule
Cluster $-$ single, $T=1.0$ & $+0.22$ & $[+0.03, +0.40]$ \\
Cluster $-$ single, $T=1.5$ & $+0.17$ & $[-0.11, +0.42]$ \\
Temperature cost (single $T{=}1.0$ $-$ single $T{=}1.5$) & $+0.81$ & $[+0.56, +1.07]$ \\
Net (cluster $T{=}1.5$ $-$ single $T{=}1.0$) & $-0.64$ & $[-0.93, -0.37]$ \\
\bottomrule
\end{tabular}
\end{table}

\subsection{When It Fails, and Where It Applies}
The method fails wherever the samples collapse, the common case on open-ended text: 73 of the
120 open-ended prompts have a largest-cluster share at or above 0.95, so failure is a property
of the regime. Voting helps only where the samples are both varied and individually good, which
meets on checkable tasks (accurate, with differing errors) but not on open-ended text from a
small instruct model, where the good answers are near-identical and the varied ones worse.

\subsection{Open Questions}
Four questions remain: whether the diversity-quality trade-off eases at larger scale, since our
scale evidence is a single 7B diversity spot-check with no quality run; whether alignment can be
separated from the competence it brings, which our base-versus-instruct comparison varies
together; whether diversity interventions other than temperature (min-$p$ sampling, diverse
decoding, prompt-induced variation) can supply diversity that stays coherent; and whether the
LLM judge, validated here against two authors ($\rho = 0.79$) and a second judge
($\rho = 0.60$), would hold against a broad human study.

\FloatBarrier

\section{Conclusion}
\label{sec:conclusion}
Test-time voting improves open-ended generation at small scale only marginally, for a reason
worth more than the method: a frozen instruct model places almost every open-ended sample in one
meaning cluster, so there is little to select over and plain and semantic voting come out the
same. Voting still wins where answers are checkable (a paired $+8.5$ math points). The collapse
tracks instruction-tuning rather than scale, and higher temperature re-opens diversity only at a
quality cost voting cannot cross. The practical reading is narrow: judge-free voting earns its
cost on tasks that already admit a cheap check, not on the open-ended tasks that motivated it.
The broader reading reframes the problem: the binding question is not how to combine samples but
whether a model produces samples worth combining, which moves it from the aggregator to the
sampler, and at small scale to what alignment does to it. Whether a richer intervention can
supply coherent diversity, and whether the trade-off eases with scale, we leave open.
\FloatBarrier

\section*{Limitations}
\paragraph{The judge is the measuring instrument.} Open-ended and long-form quality is
whatever \texttt{gpt-4o-mini} says it is. We check this rather than assume it: a second
judge from a different family, Llama-3.1-8B-Instruct, agrees on the key comparison at
Spearman $\rho = 0.60$ over 480 answers, and the sign and significance of our main
contrasts hold under both. But $\rho = 0.60$ is moderate agreement, not high, and neither
judge is validated against human preference at this scale, so we validate it directly: on a
sample of 100 open-ended answers stratified across the score range and scored blind by two
authors, the mean human rating correlates with the judge at Spearman $\rho = 0.79$
(95\% CI $[0.70, 0.85]$), as strongly as the two raters agree with each other ($\rho = 0.73$).
The agreement is on ordering, not on level: the humans rate about $1.6$ points lower on
average, so the judge's absolute scores are lenient even though its ranking tracks human
preference. This validation uses one sample with two raters and spans the full quality range
rather than the narrow band the collapsed open-ended tier occupies.
The judge is also served through an API and is not exactly reproducible across calls or
dates, which is why we release the raw judged answers rather than only the scores.

\paragraph{The diversity metric conflates form with content.} Largest-cluster share is
computed from embeddings of full responses, so a set of samples that share a template but
differ in substance is counted as collapsed. This matters most on the open-ended prompts
that happen to have a factual core, where the model can vary the answer inside a
near-identical frame. Our collapse numbers should therefore be read as an upper bound on
how concentrated the \emph{content} is.

\paragraph{The base-model comparison is confounded twice.} It varies alignment and
competence together, so it cannot attribute the diversity difference to instruction-tuning
alone. It also runs the base model as a raw completion rather than through a chat
template, so the judge may be penalising unformatted style as well as content, which makes
the $-2.23$ quality gap an upper bound on the real difference.

\paragraph{Scope.} The main voting result covers three small instruct families
(Qwen2.5-3B, Llama-3.2-3B, Gemma-2-2B) at two generation seeds, but the mechanism
analysis, the base-versus-instruct and larger-model diversity comparisons, the temperature
decomposition, and the 200-problem math run, is carried out on the primary family alone, so
the attribution to instruction-tuning is established for one family and assumed, not
retested, for the other two. The 7B scale control is a 20-prompt diversity spot-check with
no quality run. We use one judge family and one open-ended source (Dolly-15k). We test only
decoding temperature as a diversity intervention, so the negative result speaks to
temperature-based recovery and not to min-$p$ sampling, diverse decoding, or prompt-induced
variation, all of which remain open.

\paragraph{Verbose models are embedded past the encoder's window.} Clustering embeds each
open-ended answer with a 384-token encoder, and the two more verbose families exceed that
window often: 42 percent of Llama answers and 46 percent of Gemma answers are truncated
before embedding, against 6 percent for Qwen. Truncation can only remove information the
clusterer would use, so it works against cluster voting rather than for it, and the
open-ended ceiling is set by the flat single-sample baseline, which does not touch the
embedder. We ran that check directly: re-embedding the open-ended answers with an 8192-token
encoder (\texttt{bge-m3}) and repeating the selection changes the open-ended cluster-voting
score by at most $0.30$ points and $0.06$ on average across models and budgets
(Appendix~\ref{app:ablation}), so the result does not depend on the embedder's context window.

\paragraph{Answer extraction is noisier on one family.} The math control reads a boxed
final answer where the model gives one and falls back to the last number otherwise. Gemma
supplies a boxed answer on about 30 percent of its math samples, against most for Qwen, so
its math accuracy leans on the fallback and should be read as a floor; the within-model
scaling of cluster voting over the budget holds regardless, and math is a control rather
than a headline tier.

\section*{Ethics Statement}
\paragraph{What consensus selection does to minority answers.} Our aggregator returns the
representative of the largest semantic cluster. On a question with one defensible answer
this is a reasonable rule. On a question where reasonable people differ, it is a rule that
systematically discards the minority view and returns the modal one, and it does so
without any signal that a choice was made. Deployed as a quality improvement on
open-ended generation, it would narrow the range of answers a user sees. Our own results
limit the immediate concern, since the samples are already collapsed and the aggregator
changes little, but that is a statement about this model at this scale, not a property of
the method. Anyone applying consensus selection where samples are genuinely diverse
should treat the suppression of minority answers as the expected behaviour rather than an
edge case.

\paragraph{A finding about homogenisation.} We report that instruction-tuning sharply
reduces a model's output diversity, consistent with prior work. We note it here because
the effect compounds: the narrowing is in the model, and consensus selection on top of it
narrows further. Neither step is visible to a user.

\paragraph{Data.} All prompts come from public datasets used within their licences:
Dolly-15k (CC BY-SA 3.0), TruthfulQA (Apache 2.0), MT-Bench (Apache 2.0), and MATH-500, a
subset of MATH (MIT). These are curated, publicly released sets that carry no personal
data, and we did not collect, annotate, or crowdsource anything. No human subjects were
recruited; the only human evaluation is the judge-validation exercise scored by the two
authors themselves on public-dataset outputs, so no external participants or personal data
were involved.

\paragraph{Models and compute.} Generation and embedding ran on a single NVIDIA L4 GPU,
for a total of about 45 GPU-hours across the six main runs, plus a few hours each for the
primary-model diagnostics. Judging used \texttt{gpt-4o-mini} through
the OpenAI API, so benchmark prompts and model outputs were sent to a third-party service;
all of that content is public-dataset material. We released no model weights and trained
nothing.

\paragraph{Unscreened outputs.} The responses we release are raw generations from a small
instruct model on open-ended prompts. We did not filter them for harmful or inaccurate
content, and they should be treated as model output rather than as vetted text.

% ...

\bibliography{custom}

@inproceedings{choi2026modex,
    title = "{M}ode{X}: Evaluator-Free Best-of-N Selection for Open-Ended Generation",
    author = "Choi, Hyeong Kyu  and Li, Sharon",
    editor = "Liakata, Maria  and Moreira, Viviane P.  and Zhang, Jiajun  and Jurgens, David",
    booktitle = "Proceedings of the 64th Annual Meeting of the {A}ssociation for {C}omputational {L}inguistics (Volume 1: Long Papers)",
    month = jul,
    year = "2026",
    address = "San Diego, California, United States",
    publisher = "Association for Computational Linguistics",
    url = "https://aclanthology.org/2026.acl-long.655/",
    doi = "10.18653/v1/2026.acl-long.655",
    pages = "14394--14416"
}

@inproceedings{jang2026cannot,
    title = "Instruction-Tuned Language Models Cannot Sample from Distributions They Can Describe",
    author = "Jang, Chaemin  and Kim, Jihee  and Lee, Dongman",
    editor = "Lee, Hung-Yi  and Dev, Sunipa  and Li, Jessy  and Elliott, Desmond",
    booktitle = "Findings of the {A}ssociation for {C}omputational {L}inguistics: {EMNLP} 2026",
    month = oct,
    year = "2026",
    address = "Budapest, Hungary",
    publisher = "Association for Computational Linguistics",
    url = "https://aclanthology.org/2026.findings-emnlp.2278/",
    pages = "48757--48777"
}

@inproceedings{kumar2004mbr,
    title = "Minimum {B}ayes-Risk Decoding for Statistical Machine Translation",
    author = "Kumar, Shankar  and Byrne, William",
    booktitle = "Proceedings of the Human Language Technology Conference of the {N}orth {A}merican Chapter of the {A}ssociation for {C}omputational {L}inguistics: {HLT}-{NAACL} 2004",
    year = "2004",
    address = "Boston, Massachusetts, USA",
    publisher = "Association for Computational Linguistics",
    url = "https://aclanthology.org/N04-1022/",
    pages = "169--176"
}

@article{farquhar2024semantic,
  title   = {Detecting hallucinations in large language models using semantic entropy},
  author  = {Farquhar, Sebastian and Kossen, Jannik and Kuhn, Lorenz and Gal, Yarin},
  journal = {Nature},
  volume  = {630},
  pages   = {625--630},
  year    = {2024},
  doi     = {10.1038/s41586-024-07421-0}
}

@article{deepseek2025r1,
  title   = {{DeepSeek-R1} incentivizes reasoning in {LLMs} through reinforcement learning},
  author  = {{DeepSeek-AI}},
  journal = {Nature},
  year    = {2025},
  doi     = {10.1038/s41586-025-09422-z},
  url     = {https://doi.org/10.1038/s41586-025-09422-z}
}

@inproceedings{snell2024scaling,
  title     = {Scaling {LLM} Test-Time Compute Optimally Can Be More Effective than Scaling Parameters for Reasoning},
  author    = {Snell, Charlie and Lee, Jaehoon and Xu, Kelvin and Kumar, Aviral},
  booktitle = {International Conference on Learning Representations (ICLR)},
  year      = {2025},
  url       = {https://arxiv.org/abs/2408.03314}
}

@inproceedings{lightman2023verify,
  title     = {Let's Verify Step by Step},
  author    = {Lightman, Hunter and Kosaraju, Vineet and Burda, Yura and Edwards, Harrison and Baker, Bowen and Lee, Teddy and Leike, Jan and Schulman, John and Sutskever, Ilya and Cobbe, Karl},
  booktitle = {International Conference on Learning Representations (ICLR)},
  year      = {2024},
  url       = {https://arxiv.org/abs/2305.20050}
}

@inproceedings{wang2023selfconsistency,
  title     = {Self-Consistency Improves Chain of Thought Reasoning in Language Models},
  author    = {Wang, Xuezhi and Wei, Jason and Schuurmans, Dale and Le, Quoc and Chi, Ed and Narang, Sharan and Chowdhery, Aakanksha and Zhou, Denny},
  booktitle = {International Conference on Learning Representations (ICLR)},
  year      = {2023},
  url       = {https://arxiv.org/abs/2203.11171}
}

@inproceedings{kuhn2023semantic,
  title     = {Semantic Uncertainty: Linguistic Invariances for Uncertainty Estimation in Natural Language Generation},
  author    = {Kuhn, Lorenz and Gal, Yarin and Farquhar, Sebastian},
  booktitle = {International Conference on Learning Representations (ICLR)},
  year      = {2023},
  url       = {https://arxiv.org/abs/2302.09664}
}

@inproceedings{kirk2024understanding,
  title     = {Understanding the Effects of {RLHF} on {LLM} Generalisation and Diversity},
  author    = {Kirk, Robert and Mediratta, Ishita and Nalmpantis, Christoforos and Luketina, Jelena and Hambro, Eric and Grefenstette, Edward and Raileanu, Roberta},
  booktitle = {International Conference on Learning Representations (ICLR)},
  year      = {2024},
  url       = {https://arxiv.org/abs/2310.06452}
}

@inproceedings{padmakumar2024writing,
  title     = {Does Writing with Language Models Reduce Content Diversity?},
  author    = {Padmakumar, Vishakh and He, He},
  booktitle = {International Conference on Learning Representations (ICLR)},
  year      = {2024},
  url       = {https://arxiv.org/abs/2309.05196}
}

@inproceedings{holtzman2020curious,
  title     = {The Curious Case of Neural Text Degeneration},
  author    = {Holtzman, Ari and Buys, Jan and Du, Li and Forbes, Maxwell and Choi, Yejin},
  booktitle = {International Conference on Learning Representations (ICLR)},
  year      = {2020},
  url       = {https://arxiv.org/abs/1904.09751}
}

@inproceedings{nguyen2025minp,
  title     = {Turning Up the Heat: Min-p Sampling for Creative and Coherent {LLM} Outputs},
  author    = {Minh, Nguyen Nhat and Baker, Andrew and Neo, Clement and Roush, Allen and Kirsch, Andreas and Shwartz-Ziv, Ravid},
  booktitle = {International Conference on Learning Representations (ICLR)},
  year      = {2025},
  url       = {https://arxiv.org/abs/2407.01082}
}

@inproceedings{zheng2023judging,
  title     = {Judging {LLM}-as-a-Judge with {MT-Bench} and Chatbot Arena},
  author    = {Zheng, Lianmin and Chiang, Wei-Lin and Sheng, Ying and Zhuang, Siyuan and Wu, Zhanghao and Zhuang, Yonghao and Lin, Zi and Li, Zhuohan and Li, Dacheng and Xing, Eric P. and Zhang, Hao and Gonzalez, Joseph E. and Stoica, Ion},
  booktitle = {Advances in Neural Information Processing Systems (NeurIPS)},
  year      = {2023},
  url       = {https://arxiv.org/abs/2306.05685}
}

@inproceedings{ouyang2022instructgpt,
  title     = {Training Language Models to Follow Instructions with Human Feedback},
  author    = {Ouyang, Long and Wu, Jeffrey and Jiang, Xu and Almeida, Diogo and Wainwright, Carroll L. and Mishkin, Pamela and Zhang, Chong and Agarwal, Sandhini and Slama, Katarina and Ray, Alex and Schulman, John and Hilton, Jacob and Kelton, Fraser and Miller, Luke and Simens, Maddie and Askell, Amanda and Welinder, Peter and Christiano, Paul and Leike, Jan and Lowe, Ryan},
  booktitle = {Advances in Neural Information Processing Systems (NeurIPS)},
  year      = {2022},
  url       = {https://arxiv.org/abs/2203.02155}
}

@inproceedings{hendrycks2021math,
  title     = {Measuring Mathematical Problem Solving With the {MATH} Dataset},
  author    = {Hendrycks, Dan and Burns, Collin and Kadavath, Saurav and Arora, Akul and Basart, Steven and Tang, Eric and Song, Dawn and Steinhardt, Jacob},
  booktitle = {Advances in Neural Information Processing Systems Track on Datasets and Benchmarks (NeurIPS Datasets and Benchmarks)},
  year      = {2021},
  url       = {https://arxiv.org/abs/2103.03874}
}

@article{brown2024monkeys,
  title   = {Large Language Monkeys: Scaling Inference Compute with Repeated Sampling},
  author  = {Brown, Bradley and Juravsky, Jordan and Ehrlich, Ryan and Clark, Ronald and Le, Quoc V. and R{\'e}, Christopher and Mirhoseini, Azalia},
  journal = {arXiv preprint arXiv:2407.21787},
  year    = {2024},
  url     = {https://arxiv.org/abs/2407.21787}
}

@article{chen2023universal,
  title   = {Universal Self-Consistency for Large Language Model Generation},
  author  = {Chen, Xinyun and Aksitov, Renat and Alon, Uri and Ren, Jie and Xiao, Kefan and Yin, Pengcheng and Prakash, Sushant and Sutton, Charles and Wang, Xuezhi and Zhou, Denny},
  journal = {arXiv preprint arXiv:2311.17311},
  year    = {2023},
  url     = {https://arxiv.org/abs/2311.17311}
}

@article{cobbe2021verifiers,
  title   = {Training Verifiers to Solve Math Word Problems},
  author  = {Cobbe, Karl and Kosaraju, Vineet and Bavarian, Mohammad and Chen, Mark and Jun, Heewoo and Kaiser, Lukasz and Plappert, Matthias and Tworek, Jerry and Hilton, Jacob and Nakano, Reiichiro and Hesse, Christopher and Schulman, John},
  journal = {arXiv preprint arXiv:2110.14168},
  year    = {2021},
  url     = {https://arxiv.org/abs/2110.14168}
}

@article{uesato2022process,
  title   = {Solving Math Word Problems with Process- and Outcome-Based Feedback},
  author  = {Uesato, Jonathan and Kushman, Nate and Kumar, Ramana and Song, Francis and Siegel, Noah and Wang, Lisa and Creswell, Antonia and Irving, Geoffrey and Higgins, Irina},
  journal = {arXiv preprint arXiv:2211.14275},
  year    = {2022},
  url     = {https://arxiv.org/abs/2211.14275}
}

@article{bai2022hh,
  title   = {Training a Helpful and Harmless Assistant with Reinforcement Learning from Human Feedback},
  author  = {Bai, Yuntao and Jones, Andy and Ndousse, Kamal and Askell, Amanda and Chen, Anna and DasSarma, Nova and Drain, Dawn and Fort, Stanislav and Ganguli, Deep and Henighan, Tom and Joseph, Nicholas and Kadavath, Saurav and Kernion, Jackson and Conerly, Tom and El-Showk, Sheer and Elhage, Nelson and Hatfield-Dodds, Zac and Hernandez, Danny and Hume, Tristan and Johnston, Scott and Kravec, Shauna and Lovitt, Liane and Nanda, Neel and Olsson, Catherine and Amodei, Dario and Brown, Tom and Clark, Jack and McCandlish, Sam and Olah, Chris and Mann, Ben and Kaplan, Jared},
  journal = {arXiv preprint arXiv:2204.05862},
  year    = {2022},
  url     = {https://arxiv.org/abs/2204.05862}
}

@article{qwen2024,
  title   = {{Qwen2.5} Technical Report},
  author  = {{Qwen Team}},
  journal = {arXiv preprint arXiv:2412.15115},
  year    = {2024},
  url     = {https://arxiv.org/abs/2412.15115}
}

@misc{openai2024o1,
  title        = {Learning to Reason with {LLMs}},
  author       = {{OpenAI}},
  year         = {2024},
  howpublished = {\url{https://openai.com/index/learning-to-reason-with-llms/}}
}

@misc{yasser2026ttsbenchmark,
  author    = {Yasser, Adam and Elafifi, Abdullah},
  title     = {Test-Time Scaling for Open-Ended Reasoning at Small Model Scale: Benchmark and Code},
  year      = {2026},
  publisher = {Zenodo},
  doi       = {10.5281/zenodo.23669590},
  url       = {https://doi.org/10.5281/zenodo.23669590}
}

@article{dubey2024llama3,
  title   = {The {Llama} 3 Herd of Models},
  author  = {{Llama Team, AI @ Meta}},
  journal = {arXiv preprint arXiv:2407.21783},
  year    = {2024},
  url     = {https://arxiv.org/abs/2407.21783}
}

@article{gemma2024,
  title   = {{Gemma} 2: Improving Open Language Models at a Practical Size},
  author  = {{Gemma Team}},
  journal = {arXiv preprint arXiv:2408.00118},
  year    = {2024},
  url     = {https://arxiv.org/abs/2408.00118}
}

\appendix
\section{Implementation Details}
\label{app:details}

\paragraph{Selection procedure.} Algorithm~\ref{alg:method} states the full test-time selection
procedure: the shared sampling step and the three selectors.

\begin{algorithm}[!ht]
\caption{Test-time selection: shared sampling, three selectors.}
\label{alg:method}
\begin{algorithmic}[1]
\Require prompt $x$, budget $N$, frozen model $M$, embedder $\phi$, method $\in$ \{\textsc{single}, \textsc{usc}, \textsc{cluster}\}
\State sample $y_1, \dots, y_N \sim M(\cdot \mid x)$ once, kept in generation order \Comment{shared by all}
\If{method $=$ \textsc{single}}
  \State \Return $y_1$
\ElsIf{method $=$ \textsc{usc}}
  \State $k \gets$ model's parsed choice of the most consistent sample (first sample if unparsable)
  \State \Return $y_k$
\Else \Comment{semantic cluster voting}
  \For{$i = 1$ \textbf{to} $N$}
    \State $u_i \gets$ extracted answer of $y_i$ if tier is math, else $y_i$
    \State $e_i \gets \phi(u_i) / \lVert \phi(u_i) \rVert$
  \EndFor
  \State $D_{ij} \gets \max(0,\, 1 - e_i \cdot e_j)$ for all $i, j$; set the diagonal to $0$
  \State build the average-linkage dendrogram and cut at $\tau$ (largest merge gap)
  \State $C^\star \gets$ largest cluster (size ties to the lower label)
  \If{tier $=$ \textsc{math}}
    \State \Return plurality of $\{\,\text{extracted answer of } y_i : i \in C^\star\,\}$ (ties to earliest sample)
  \Else
    \State $m^\star \gets \arg\max_{i \in C^\star} \sum_{j \in C^\star} e_i \cdot e_j$
    \State \Return $y_{m^\star}$
  \EndIf
\EndIf
\end{algorithmic}
\end{algorithm}

\paragraph{Math answer extraction.} We take a math answer in two steps. First we
read the contents of the last \texttt{\textbackslash boxed\{...\}}, matching nested
braces. When no boxed expression is present, we fall back to the last number in the
response, found with the pattern \texttt{-?\textbackslash d+(\textbackslash.\textbackslash d+)?}
after removing thousands separators. We normalize the extracted string by stripping
surrounding whitespace and a leading currency sign, removing commas, spaces, and the
thin-space \texttt{\textbackslash !}, and dropping a leading plus sign; an empty
result counts as no answer. Scoring on the math tier is exact match between the
normalized prediction and the normalized reference.

\paragraph{Universal self-consistency prompt.} For USC we show the model all $N$
candidates, each truncated to 800 characters so the full set fits in context, and
number them from one. The instruction reads: ``Below are $N$ candidate answers to
the same question. Pick the ONE that is most consistent with the majority of the
others. Reply with ONLY its number.'' We append the question and the numbered
candidates, decode the reply greedily with a short generation budget, and read the
first integer in the reply as the chosen index. When the reply holds no valid index,
or when $N = 1$, we return the first candidate.

\paragraph{Generation.} Prompts are wrapped in the Qwen chat template; math prompts
additionally receive the instruction ``Solve step by step and put your final answer in
\textbackslash boxed\{\}''. We generate at temperature 1.0 and top-$p$ 0.95 under seed
42, in batches of 16, with a per-tier cap on new tokens of 1024 for the text tiers and
2048 for math.

\paragraph{Diversity measurements.} The diversity analysis (Table~\ref{tab:diversity})
draws 8 samples per prompt with a 512-token cap. The temperature sweep pairs each
temperature with a top-$p$ value: 0.95 for $T \le 1.3$ and 0.98 for $T \in \{1.5,
1.7\}$. The base model is prompted as a plain completion rather than through the chat
template. The base-versus-instruct comparison and the temperature sweep use all 120
open-ended prompts with bootstrap medians over 2000 resamples; the 7B scale control is
a 20-prompt spot-check over the first 20 open-ended prompts. All confidence intervals in
the paper use 2000 bootstrap resamples. Remaining settings (library versions and random
states) are in the released configuration.

\section{Ablations}
\label{app:ablation}

\paragraph{Robustness to the clustering cut.} The cut rule is untuned, so we
repeat the central measurements under a fixed cosine-distance threshold of 0.15.
On the open-ended tier the largest-cluster share is 1.00 (95\% CI $[0.98, 1.00]$)
under the fixed cut and 0.97 ($[0.95, 0.97]$) under the adaptive cut, so the
collapse is a property of the model rather than of the cut, and the adaptive rule
if anything reports slightly more diversity. The cluster-voting scores track the
main numbers: open-ended 7.93 against 7.93, long-form QA 6.38 against 6.44, and
math 72 against 74 percent. The adaptive cut is marginally better on the
verifiable tier and does not drive the findings.

\paragraph{Clustering versus plain voting.} We ask whether the clustering step
adds anything over plain aggregation. On math we compare cluster voting to a plain
majority vote over all $N$ answers; on the text tiers we compare the cluster medoid
to the global medoid over all $N$ samples. The two are within measurement noise on
every tier: math 74 against 74 percent, open-ended 7.91 against 7.93, and long-form
QA 6.44 against 6.44. This follows directly from the collapse. When almost all
samples fall into one cluster, the largest cluster is effectively the whole sample
set, so cluster voting reduces to plain voting. The clustering neither helps nor
hurts at this scale, which is why we present it as an instrument for reading
diversity rather than as a contribution in its own right.

\paragraph{A judge-free baseline.} We check cluster voting against the cheapest
judge-free selector, returning the longest response. Across tiers the
longest-answer rule matches the single-sample floor exactly (math 64 percent,
open-ended 7.69, long-form QA 6.26) and never reaches cluster voting (74, 7.93,
6.44). Choosing the longest answer buys nothing over no selection at all, while
cluster voting improves on both. A fuller sweep of judge-free selectors, including
perplexity-based selection, remains future work.

\paragraph{Embedder.} Swapping the primary embedder for all-MiniLM-L6-v2 leaves the
results unchanged: open-ended 7.93 against 7.93, long-form QA 6.54 against 6.44, and math
74 against 74 percent. We also address the context window directly, since the more verbose
models exceed the primary encoder's 384-token limit on many open-ended answers (42 percent
for Llama, 46 percent for Gemma). Re-embedding the open-ended answers of the three seed-42
runs with an 8192-token encoder (\texttt{bge-m3}) and repeating the selection changes the
cluster-voting score by at most 0.30 points and 0.06 on average across all budgets and
models, with mixed sign and the open-ended ceiling unchanged. The finding does not depend on
the embedding model or on its context window.

\paragraph{USC and candidate truncation.} Because USC must hold all $N$ candidates in
one context, the main run caps each candidate at 800 characters. Rerunning USC on the
open-ended tier with a 4000-character budget leaves it unchanged (within 0.2 of a point
at every $N$) and does not lift it above the single-sample floor, so the cap does not
account for USC's behavior.

\paragraph{Math power run.} The 50-problem math control in the body is confirmed on 200
MATH-500 problems. Cluster voting raises accuracy from $59\%$ (single) to $67.5\%$ at $N=32$, a
paired $+8.5$ points whose 95\% bootstrap CI excludes zero at every $N \ge 4$
(Table~\ref{tab:mathpower}).

\begin{table}[htbp]
\caption{Math power run: single versus cluster voting on 200 MATH-500 problems (accuracy
\%), with the paired gain and its 95\% bootstrap CI. The gain excludes zero at every
$N \ge 4$.}
\label{tab:mathpower}
\centering
\small
\begin{tabular}{rccc}
\toprule
$N$ & Single & Cluster & Cluster $-$ Single \\
\midrule
1  & 59.0 & 59.0 & $+0.0\ [+0.0, +0.0]$ \\
4  & 59.0 & 63.0 & $+4.0\ [+1.0, +7.5]$ \\
8  & 59.0 & 66.0 & $+7.0\ [+3.0, +11.5]$ \\
16 & 59.0 & 67.0 & $+8.0\ [+3.5, +12.5]$ \\
32 & 59.0 & 67.5 & $+8.5\ [+4.0, +13.0]$ \\
\bottomrule
\end{tabular}
\end{table}

\paragraph{Base-model diagnostic.} On a separate 40-prompt open-ended set, the pre-alignment
base model is more diverse than the instruct model (largest-cluster share 0.75 against 1.00) but
far lower quality, and voting over it does not reach the instruct single-sample floor
(Table~\ref{tab:basediag}).

\begin{table}[htbp]
\caption{Base-model diagnostic on 40 open-ended prompts (8 samples each), judged 1 to 10 with
95\% bootstrap CIs. The base model is more diverse (largest-cluster share 0.75 versus the
instruct model's 1.00) but far lower quality, and voting over it does not reach the instruct
single-sample floor.}
\label{tab:basediag}
\centering
\begin{tabular}{lcc}
\toprule
Condition & Quality /10 & 95\% CI \\
\midrule
Base, single     & 5.53 & $[4.67, 6.35]$ \\
Base, cluster    & 6.20 & $[5.38, 6.95]$ \\
Instruct, single & 7.75 & $[7.08, 8.35]$ \\
\bottomrule
\end{tabular}
\end{table}

\section{Per-Budget Paired Intervals}
\label{app:intervals}
Table~\ref{tab:intervals} reports the paired difference over prompts, with 95\% bootstrap
confidence intervals, for cluster voting and USC against the single-sample baseline at
every sample budget $N$. A positive value favors the method; an interval that includes
zero is not distinguishable from the baseline. Text tiers are in judge points; the math
tier is in accuracy points on the 50-problem set.

\begin{table}[htbp]
\caption{Paired differences against single (95\% bootstrap CI). Text in judge points, math
in accuracy points on the 50-problem tier.}
\label{tab:intervals}
\centering
\small
\begin{tabular}{rll}
\toprule
$N$ & Cluster $-$ Single & USC $-$ Single \\
\midrule
\multicolumn{3}{l}{\emph{Open-ended}} \\
1  & $+0.00\ [+0.00, +0.00]$ & $+0.00\ [+0.00, +0.00]$ \\
4  & $+0.03\ [-0.14, +0.19]$ & $+0.10\ [-0.06, +0.28]$ \\
8  & $+0.13\ [-0.03, +0.31]$ & $+0.33\ [+0.15, +0.53]$ \\
16 & $+0.12\ [-0.07, +0.31]$ & $+0.21\ [+0.04, +0.40]$ \\
32 & $+0.15\ [-0.03, +0.35]$ & $+0.05\ [-0.13, +0.23]$ \\
64 & $+0.24\ [+0.05, +0.44]$ & $+0.00\ [-0.18, +0.20]$ \\
\midrule
\multicolumn{3}{l}{\emph{Long-form QA}} \\
1  & $+0.00\ [+0.00, +0.00]$ & $+0.00\ [+0.00, +0.00]$ \\
4  & $+0.04\ [-0.35, +0.41]$ & $+0.11\ [-0.26, +0.49]$ \\
8  & $+0.10\ [-0.26, +0.44]$ & $+0.10\ [-0.29, +0.49]$ \\
16 & $+0.09\ [-0.33, +0.51]$ & $+0.05\ [-0.33, +0.38]$ \\
32 & $+0.17\ [-0.26, +0.60]$ & $+0.01\ [-0.36, +0.39]$ \\
64 & $+0.09\ [-0.36, +0.54]$ & $-0.17\ [-0.70, +0.34]$ \\
\midrule
\multicolumn{3}{l}{\emph{Math (50 problems)}} \\
1  & $+0\ [+0, +0]$   & $+0\ [+0, +0]$ \\
4  & $+0\ [-6, +6]$   & $-6\ [-16, +4]$ \\
8  & $+0\ [-10, +10]$ & $-4\ [-14, +6]$ \\
16 & $+4\ [-4, +12]$  & $-12\ [-24, -2]$ \\
32 & $+10\ [0, +20]$  & $+0\ [-14, +12]$ \\
64 & $+6\ [-2, +14]$  & $+0\ [-8, +8]$ \\
\bottomrule
\end{tabular}
\end{table}

\end{document}